\documentclass{article}
\usepackage{gb4e}
\usepackage{qtree}

\makeatletter
\def\literaturename{References}

\def\thebibliography#1{\section*{\literaturename}\small\list
  {\arabic{enumi}.}{\settowidth\labelwidth{#1.}\leftmargin\labelwidth
    \advance\leftmargin\labelsep
    \usecounter{enumi}}
    \def\newblock{\hskip .11em plus .33em minus -.07em}
    \sloppy
    \sfcode`\.=1000\relax}

\def\ds@citeauthoryear{\def\thebibliography##1{\section*{\literaturename}%
    \small\list{}{\settowidth\labelwidth{}\leftmargin\parindent
    \itemindent=-\parindent
    \labelsep=\z@
    \usecounter{enumi}}%
    \def\newblock{\hskip .11em plus .33em minus -.07em}%
    \sloppy
    \sfcode`\.=1000\relax}%
    \def\@cite##1{##1}%
    \def\@lbibitem[##1]##2{\item[]\if@filesw
      {\def\protect####1{\string ####1\space}\immediate
    \write\@auxout{\string\bibcite{##2}{##1}}}\fi\ignorespaces}}%

\let\@internalcite\cite
\def\cite{\def\citename##1{##1, }\@internalcite}
\def\shortcite{\def\citename##1{}\@internalcite}
\def\newcite{\leavevmode\def\citename##1{{##1} (}\@internalciteb}

\skip\footins 9pt plus 4pt minus 2pt
\@beginparpenalty-\@lowpenalty\@endparpenalty -\@lowpenalty\@itempenalty
-\@lowpenalty

\def\section{\@startsection {section}{1}{\z@}{-10pt plus -4pt minus
-4pt}{9pt plus 4pt minus 4pt}{\large\bf
\pretolerance=10000\relax\rightskip=0pt plus8em}}
\def\subsection{\@startsection{subsection}{2}{\z@}{-10pt plus-4pt minus
 -4pt}{8pt plus 4pt minus 4pt}{\normalsize\bf\boldmath
\pretolerance=10000\relax\rightskip=0pt plus8em}}
\def\subsubsection{\@startsection{subsubsection}{3}{\z@}{-10pt plus-4pt
 minus -4pt}{-0.5em plus -.22em minus -0.1em}{\normalsize\bf\boldmath}}
\def\paragraph{\@startsection{paragraph}{4}{\z@}{-10pt plus -4pt minus
 -4pt}{-0.5em plus -.22em minus -0.1em}{\normalsize\it}}
\def\subparagraph#1{\typeout{LLNCS Warning: You should not use
\protect\subparagraph \space in this style.}\vskip0.5cm
You should not use $\backslash${\tt subparagraph} in this
style.\vskip0.5cm}

\makeatother

\singlegloss
\newcommand{\rsent}[1]{(\ref{sent:#1})}

\newcommand{\jpn}[2]{{\it #2} ``#1''}
\newcommand{\iz}[1]{{\tt #1}}
\newcommand{\eng}[1]{{\it #1\/}}
\newcommand{\engs}[1]{{\it #1}}
\newcommand{\ul}{\underline}
\newcommand{\into}{$\Rightarrow$\ }

\newcommand{\com}[1]{\hfill (#1)}

\newcommand{\border}{\vspace*{-7mm}%
\rule{\textwidth}{0.2mm}
\vspace*{-8mm}}
\renewcommand{\newcite}{\cite}

\title{\vspace*{-20mm}Temporal Expressions\\ in Japanese-to-English Machine Translation}
\author{Francis Bond, Kentaro Ogura and Hajime Uchino}
\date{{\bf NTT Communication Science Laboratories} \\ 
  1-1 Hikari-no-oka, Yokosuka-shi, Kanagawa-ken, {\sc Japan} 239 \\ 
  {\tt \{bond,ogura,uchino\}@cslab.kecl.ntt.co.jp}\\[2mm]
TMI-97, Santa Fe, July 1997, pp 55--62\\[2mm]
\textbf{Abstract} \small
  \begin{quotation}
  This paper describes in outline a method for
  translating Japanese temporal expressions into English. We argue
  that temporal expressions form a special subset of language that is
  best handled as a special module in machine translation.  The paper
  deals with problems of lexical idiosyncrasy as well as the choice of
  articles and prepositions within temporal expressions.  In addition
  temporal expressions are considered as parts of larger structures,
  and the question of whether to translate them as noun phrases or
  adverbials is addressed.
  \end{quotation}
}

\begin{document}
\maketitle


\vspace*{-10mm}\section{Introduction}
\label{sec:intro}

In this paper we argue that the transfer and generation of temporal
noun phrases and adverbials are best handled by a separate (though
integrated) module in Japanese-to-English machine translation.

There are several reasons for creating a separate module.  They center
on the fact that temporal expressions are lexically and syntactically
highly idiosyncratic in both Japanese and English.  In addition to
grammatical differences, there are differences in style between the
two language communities that need to be addressed by a machine
translation system: for example in English stock market reports days
are given as days of the week, whereas in Japanese they are given as
dates.

We argue that attempting to handle all phenomena by a single
mechanism, while conceptually elegant, causes unnecessary
complications.  In addition trying to shoehorn all idiosyncrasies into
any one method is likely to strain it to the breaking point.  Instead
we adopt the multi-level machine translation approach of
\newcite{Ikehara:1991} in which there are many levels of transfer
between two languages, and expressions are transferred at whatever
level the system judges to be the most appropriate.  

The processing described has been implemented in the
Japanese-to-English machine translation system \textbf{ALT-J/E}
\cite{Ikehara:1991,Ikehara:1996b}.

\section{Temporal Noun Phrases}
\label{sec:NP}

English temporal noun phrases, along with locatives, do not have all
the properties of prototypical `purebred' noun phrases. They are one
of the five types of noun phrases classed as `defective' by
\newcite{Ross:1995}.  For example, in contrast to `purebred' noun
phrases, they are typically pronominalized by \eng{then} rather than
\engs{it}.  There is also a great deal of variation in realization
between dialects, with substantial differences between Australian,
American and British English.\footnote{There is also some dialectal
  variation in Japanese temporal expressions.  For example, {\it
    yanoasatte}, which means the fourth day after today in the
  standard dialect, means three days after today in some areas.}
In addition, there is a lot of lexical idiosyncrasy as we will show in
the following sections.

In the following section we will restrict our discussion to noun
phrases that show position in time, rather than duration or frequency.
English temporal adjuncts are described in some detail in
\newcite[526--555]{Quirk:1985}.  To the best of our knowledge, the
only description of their treatment in natural language processing is
that of \newcite{Flickinger:1996}, who discusses English time
expressions in the HPSG grammar used by the Verbmobil
German/Japanese-to-English machine translation project.

\subsection{Temporal Noun Phrase Structure}
\label{sec:eng}

English time position noun phrases used primarily to refer to time
(for example within adverbials or the subject of sentences such as
\eng{\ul{Spring} has come}) are highly idiosyncratic in their lexical
choice, as well as their choice of determiners, typically having no
surface determiner, although some take the definite article.  Note
that noun phrases headed by the same nouns, but not primarily referring
to time, behave as do other nouns: \engs{It was \ul{a spring} to
  remember}.  To handle these lexical and syntactic idiosyncrasies we
introduce special processing for temporal noun phrases.

\newcite[433]{Ross:1995} claims that such defective noun phrases are
always locally triggered, that is there is some clause mate that
forces or enables the noun phrase to become defective.  If we allow
time position adjuncts to license themselves as being defective, then
this claim holds, and we can always count on there being some trigger
to introduce our special processing.

We will now give some example of the idiosyncrasies, starting with the
lack of an article for unmodified nouns, assuming they are in temporal
noun phrases: \rsent{null}.\footnote{After each example in
  \rsent{null} we give the semantic attribute of its head, from the
  hierarchy given in Figure~\ref{fig:hpsg}.}

\begin{exe}
  \ex \label{sent:null}
  \begin{xlist}
    \ex today, yesterday, tomorrow \com{\iz{deictic-day}}
  \ex Monday \com{\iz{day-of-week}}
  \ex Christmas \com{\iz{holiday}}
  \ex 3 o'clock, 12:15 \com{\iz{numbered-hour}} 
  \ex February \com{\iz{month}}
  \ex 1997 \com{\iz{year}}
  \ex dawn \com{\iz{time-of-day}}
  \ex winter \com{\iz{season}}
  \end{xlist}
\end{exe}

We analyze the noun phrases with no surface determiner as {\sc null}
determiners, phonologically empty determiners which appear in noun
phrases with definite reference to a locatable, one-member referent
set itself, following \newcite[73]{Chesterman:1991}.\footnote{Null
  determiners are distinct from {\sc zero} determiners, also
  phonologically empty, which appear with indefinite uncountable and
  plural noun phrases, the equivalent to \eng{a} for singular
  countable noun phrases.}  Our analysis of null determiners is
independently motivated, we also consider them to appear in proper
names, technical terms, and noun phrases such as \eng{school} in
\eng{I went to school today}.

In contrast with the noun types shown in \rsent{null}, ordinal numbers
denoting the day of a month, on the other hand, normally take the
definite article: \engs{the 19th}.

The choice of dependent also affects the choice of determiner, for
example, there are some temporal expressions which take the null
determiner when they are modified by another class of time expressions
\rsent{mod-null}:

\begin{exe}
  \ex \label{sent:mod-null}
  \begin{xlist}
    \ex Monday morning
    \ex yesterday morning
    \ex Monday night
    \ex February 19
    \ex February 19th
  \end{xlist}
\end{exe}

We analyze these simply as ``{\sc null} modifier head''.  In our
translation module, we generate the determiner during the transfer
from Japanese.  \newcite{Flickinger:1996} introduces a more elegant
analysis where the first element is treated as specifier (determiner)
of the second, thus explaining the lack of article.  Both the
specifier's change in part of speech, and the head's choice of
complement, are expressed by lexical rules.  However, as null
determiners or their equivalent are needed for the noun phrases in
\rsent{null} anyway, we do not consider their use here to be
problematic.

In addition, as \newcite{Flickinger:1996} notes, his analysis leads to
noun phrases with multiple specifiers in expressions like
\engs{February the 19th}, although most analyses of English allow only
one specifier.


To handle such cases, we use a special structure, the special compound
noun phrase, that we have established for noun phrases where there is
no obvious head, such as person and company names, and addresses.  We
use this structure for time position noun phrases which include the
following elements: \iz{year}, \iz{day-of-month}, \iz{month} and/or
\iz{numbered-time}.  In these noun phrases there is no obvious
semantic head and there are many possible representations in English.
We show some examples of the choice of expressions for a single date
in \rsent{SC-NP}.  The choice of representation is mainly a question
of style.  In particular, it does depend on the Japanese source noun
phrase, which has only two possible forms: \jpn{2 month 19
  day}{2-gatsu-19-nichi} and \jpn{2 month {\sc gen} 19
  day}{2-gatsu-no 19-nichi}.

\begin{exe}
  \ex  \label{sent:SC-NP} 
  February the 19th {\it vs\/} February 19 {\it vs\/}  
  February 19th {\it vs\/} the 19th of February
\end{exe}

By establishing a set of special structures for noun phrases that
behave atypically, and thus have to be treated atypically anyway, we
are able to preserve a uniform structure for all other noun phrases
(with a single, although potentially phonologically empty, specifier).
Although the grammar is consistent with `purebred' noun phrases, the
choice of determiner is not, which is why we argue for a separate module.

In the next section we will describe the transfer module for temporal
noun phrases, which does most of the hard work in creating appropriate 
English structures, handling lexical and phrasal idiosyncrasy.

\subsection{Transfer and Generation of Temporal Noun Phrases}
\label{sec:trans}

Temporal noun phrases in Japanese can be considered to be of three
types; those headed by single nouns, those headed by compound nouns,
and those made of one temporal noun phrase modifying another.

The transfer stage for noun phrases headed by single nouns is
basically a process of replacing them by their equivalents in the
lexicon, which may be a single English noun: \jpn{yesterday}{kin\=o} or
a phrase: \jpn{the day before yesterday}{ototoi}.  

Dates (years, months, days of months, and numbered times) are compound
nouns in Japanese, typically consisting of a number and a temporal
noun.  The compound noun rules first distinguish between time positions
and time periods, for example, the adverbial \jpn{13 days}{13-nichi}
could be \eng{on the thirteenth} or \engs{for thirteen days}.  Once it has
been determined that the noun phrase refers to time position, simple
regular rules are used to generate the corresponding English
expressions. For example, days of months in Japanese have the form
\jpn{{\sc numeral} day}{{\sc numeral}-nichi} and are translated into
special compound noun phrases with the \iz{day-of-month} slot filled
by the value for the numeral.\footnote{This is further
  complicated by the fact that the Japanese have two counting systems for
  years, one based on the western system (A.D.) and the other on years
  of the current emperor's reign.  \jpn{3 years}{3-nen} is thus
  multiply ambiguous between, at least, 3 AD, 1903 AD, 2003 AD, 1991
  AD (the third year of the current emperor's reign) and a three year
  period.  Disambiguating these readings is outside of the scope of
  this paper.}

Complex noun phrases require more complicated rules.  We show the
rules for the combination of a noun phrase denoting \iz{deictic-day}
or \iz{day} with one denoting \iz{period-of-day} (\engs{morning,
  afternoon, evening}) or \engs{night} in Figure~\ref{fig:day-period}.
The Japanese will be of the form \jpn{$\alpha$-{\sc gen}
  $\beta$}{$\alpha$-no $\beta$} where $\alpha$ is headed by a \iz{day}
or \iz{deictic-day} noun and $\beta$ is headed by a \iz{period-of-day}
noun or \eng{night},  that translates into English noun phrase
B.

\begin{figure*}[htbp]
\border
\vspace*{-5mm}
  \begin{center}
    \leavevmode
    \begin{tabbing}
B is \iz{period-of-day} (\engs{morning, afternoon, evening}) or \eng{night}:\\
if \=A is \iz{deictic-day}\\
\> if \= A = \jpn{the day before the day before yesterday}{issakusakujitsu}\\
\>\>    \into \eng{the B before the B before last}   ; 3 before\\
\>  if A = \jpn{the day before yesterday}{ototoi}\\
\>\>    \into \eng{the B before last}                ; 2 before \\
\>  if A = \jpn{yesterday}{kin\a=o} \\
\>\>    if \= B = \eng{night} \\
\>\>\>      \into  {\sc null} \eng{last night}\\
\>\>    else\\
\>\>\>    \into {\sc null} yesterday B                      ; 1 before\\
\>  if A = \jpn{today}{ky\a=o} or \jpn{today}{honjitsu} \\
\>\>     if B = \eng{night} \\
\>\>\>        \into {\sc null} \eng{tonight} \\
\>\>     else \\
\>\>\>        \into \eng{this B}                    ; today's\\
\>  if A = \jpn{tomorrow}{ashita}\\
\>\>    \into {\sc null} \eng{tomorrow B}                       ; 1 after\\
\>  if A = \jpn{the day after tomorrow}{asatte}\\
\>\>    \into \eng{the B after next}                 ; 2 after \\
\>  if A = \jpn{the day after the day after tomorrow}{shiasatte}\\
\>\>    \into \eng{the B after the B after next}     ; 3 after\\
\>  if A = \jpn{the day after the day after the day after tomorrow}{yanoasatte}\\
\>\>    \into \eng{the B after the B after the B after next}  ; 4 after  [rare]\\
if A is \iz{relative-day}\\
\>  if A = \jpn{the previous day}{zenjistu}\\
\>\>    \into \eng{the previous B}  \\
\>  if A = \jpn{the following day}{yokujitsu}\\
\>\>    \into \eng{the following B} \\
if A is \iz{named-day} \\
\>  \into {\sc null}\eng{A B}\\
if A is \iz{day-of-month}\\
\>\>  \into \eng{the B of the A} (A must be ordinal)
\end{tabbing}
      \caption{Rules for \iz{day}, \iz{deictic-day} and \iz{period-of-day}}
    \label{fig:day-period}
  \end{center}
\vspace*{3mm}
\border
\end{figure*}

These transfer rules capture the lexical and phrasal idiosyncrasies of
the temporal noun phrases, and are relatively easy to test and expand.
The rules are all specific to temporal noun phrases; no generality has
been lost by putting them in a separate module.

Note that these rules do not necessarily preserve the Japanese
structure.  In particular many temporal expressions made up of two
noun phrases in Japanese are often most naturally translated as a noun
phrase and an adverbial phrase in English.  For example,
\jpn{tomorrow-{\sc gen} dawn}{ashita-no akegata} could possibly be
translated as \engs{tomorrow's dawn}, but is generally translated as
\eng{dawn tomorrow} where \eng{tomorrow} is an adverbial modifying
\engs{dawn}.

Attempts to preserve structure, translating all Japanese temporal
noun phrases as English noun phrases, are problematic as the resulting
expressions are often unwieldy or ambiguous.  Consider for example
\jpn{this year-{\sc gen} Christmas}{kotoshi-no kurisumasu}, which
could potentially be translated as \eng{this year's Christmas},
\eng{this Christmas} or \eng{Christmas this year}.  \eng{this year's
  Christmas} sounds extremely unnatural.  The use of deictic dependents
like \eng{this} and \engs{last}, are not bound by year boundaries;
\eng{this Christmas} spoken in January, could refer to Christmas last
year, or the coming Christmas and is thus ambiguous in a way that the
Japanese original is not.\footnote{We are indebted to Tim Baldwin for reminding us of this.}
In addition, there is considerable dialectal difference in the use of
\eng{this} and \engs{next}.  Thus the most acceptable translation of
\jpn{this year-{\sc gen} Christmas}{kotoshi-no kurisumasu} is
\engs{Christmas this year}, where \eng{this year} is an adverbial.

We also found, in a preliminary investigation of bilingual corpora,
that translation of Japanese temporal noun phrases into English
adverbials was very common, not only in complex temporal noun phrases,
but also in noun phrases with only one part being temporal, for
example those in \rsent{adv-np}.

\begin{exe}
  \ex \label{sent:adv-np}
\begin{xlist}
  \ex \jpn{this week-{\sc gen} meeting}{konsh\=u-no uchiawase} 
  \trans \into this week's meeting {\it vs\/} the meeting this week
  \ex \jpn{Monday-{\sc gen} meeting}{getsuy\=obi-no uchiawase} 
   \trans \into   Monday's meeting {\it vs\/} the meeting on Monday.
\end{xlist}
\end{exe}

At present our rules translate most Japanese complex temporal noun
phrases, and all temporal noun phrases occurring inside larger
non-temporal noun phrases as noun phrases in English, but as we
identify more criteria for choosing between adverbials and noun
phrases they will be added to our module.

Note that we have similar sets of rules to those presented in
Figure~\ref{fig:day-period} for other time combinations, such as:
\engs{the January before last} (\iz{year + month}), \engs{next
  Saturday} (\iz{week + day}) and so on.

\section{Temporal Adverbials}
\label{sec:AdvP}

Japanese temporal adverbials have the same structure as noun phrases,
that is any number of modifiers, followed by a noun head, followed
by a post-positional particle.  The fact that the constituent is an
adverbial is determined by the choice of postpositional particle, the
semantic attribute of the head noun, and the structure of the
sentence.  As well as translating Japanese adverbials as adverbials it
is often necessary to translate Japanese temporal phrases into English
adverbials, as discussed in Section~\ref{sec:trans}.  In such cases,
we need to select which preposition to use.  In the next section we
outline our algorithm for choosing prepositions, a sub-section of the
module for translating temporal expressions.

\subsection{Choice of Preposition}
\label{sec:prep}

\begin{figure*}[tbp]
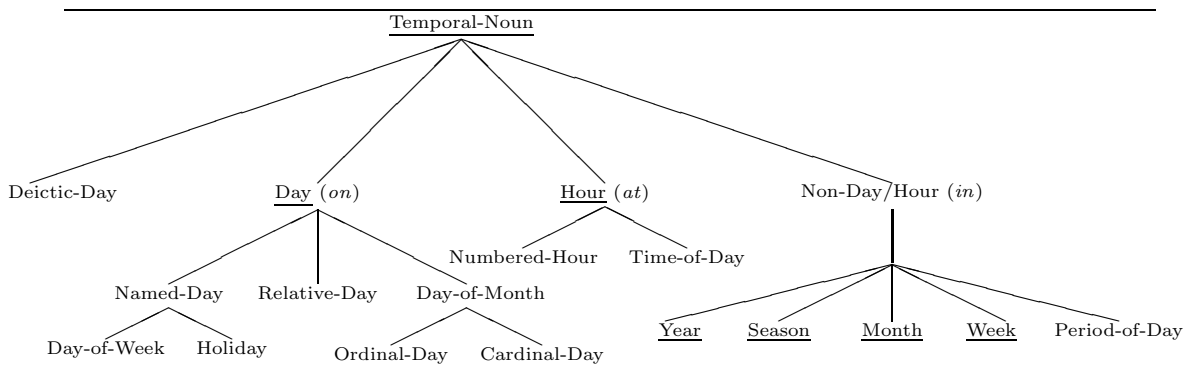

\border
\vspace*{15mm} \hspace*{-44mm}\scriptsize \Tree [ 
   {\ \ \ \ \ \ \ \ \ Deictic-Day}  !\faketreewidth{W} 
   [ 
     [ Day-of-Week Holiday ].Named-Day !\faketreewidth{WWWWWWWW}
     Relative-Day !\faketreewidth{WWW} 
     [ Ordinal-Day Cardinal-Day 
     ].{\ \ \ \ Day-of-Month} !\faketreewidth{WWWWWWWW}
   ].\ul{Day}~(\eng{on}) !\faketreewidth{WWWWWWWWWWWWWWWWWWWWWW} 
     [ Numbered-Hour Time-of-Day 
   ].\ul{Hour}~(\eng{at}) !\faketreewidth{WW} 
     [ {\ \ \ \ \ul{Year}}
     \ul{Season} 
     {\ul{Month}\ } 
     {\ul{Week}\ \ \ \ \ } 
     Period-of-Day !\faketreewidth{WWWW}
   ].\shortstack{Non-Day/Hour~(\eng{in})\\ 
     \rule[-3mm]{0.5pt}{7mm}} !\faketreewidth{W}
  ].\ul{Temporal-Noun}
\normalsize \vspace*{-5mm}
    \caption{Noun Type Hierarchy}
    \label{fig:hpsg}
\vspace*{5mm}
\border
\end{figure*}

The main problem for generating English prepositions is deciding
between \engs{on}, \eng{at} and \eng{in}.  Other prepositions, such as
\eng{before} or \eng{during} have Japanese equivalents (normally
functional nouns or post-positional particles) which can be used to
select them.  The two most widely used particles in Japanese for
temporal (and locative) expressions are {\it ni} and, to a lesser
extent, {\it de}, either of which can be translated as \engs{on},
\eng{at} or \eng{in}.  To choose between these three alternatives, we
need to consider both the semantic attribute of the head noun and its
dependents in English.  We use the semantic attribute system for
Japanese analysis of \newcite{Ikehara:1997x}, augmented with several
types defined especially in English generation.  The relevant parts of
the hierarchy are shown in Figure~\ref{fig:hpsg}.  The nodes
underlined are those that appear in the Japanese semantic attribute
system.

Our type hierarchy matches that suggested by
\newcite{Flickinger:1996}, and we adopt his nomenclature.  The
hierarchy presented here is very similar, but includes
\iz{deictic-day} nouns and more branches under the \iz{day} and
\iz{non-day/hour} node.  The new nodes are needed in writing the rules
to translate complex temporal noun phrases outlined in
Section~\ref{sec:trans}, and in choosing prepositions, as will be
shown below.

The hierarchy is loosely based on the choice of preposition for
temporal adverbials formed by a prepositional phrase followed by an
unmodified noun phrase headed by each type.  \iz{deictic-day} nouns
take no preposition, they form adverbials by themselves.

Noun phrases headed by \iz{day} nouns will take the preposition
\engs{on}.  The preposition is optional for noun phrases headed by
\iz{day-of-week} nouns in American English.  Noun phrases headed by
\iz{day-of-month} nouns can take months as specifiers in Flickinger's
analysis, \iz{cardinal-day} nouns can only take months, while
\iz{ordinal-day} nouns can take months or \eng{the} or both.  

Noun phrases headed by \iz{hour} nouns, including \iz{numbered-hour}
nouns and \iz{time-of-day} nouns like \eng{noon} and \eng{dawn} are
selected by the preposition \eng{at}.  The common denominator seems to
be the idea of a precise moment in time.

The remainder of noun phrases headed by  temporal nouns are
selected by the preposition \eng{in}.

Our algorithm for choosing whether to use a preposition, and if so
which preposition to use, is given in
Figure~\ref{fig:prep}.\footnote{In this algorithm, we treat the
  narrowest time unit as the head of temporal special compound noun
  phrases: for example, the heads of \eng{February the 19th} and
  \eng{the 19th of February} will both be \eng{19th} a
  \iz{cardinal-day} noun.  For other purposes, such as pluralization,
  the rightmost noun is treated as the head.}  The algorithm is used
to convert noun phrases headed by temporal nouns to adverbials, and
has been extended to include noun phrases of the form A \eng{of} B
where A is \eng{beginning, middle, end} and B is a temporal noun
phrase.  There are also some event nouns that could be analyzed as
heading temporal noun phrases that can appear in temporal adverbials,
such as \eng{opening} in \engs{at the opening [of the Osaka Stock
  Exchange]}, we would like to extend our analysis to include them.

\begin{figure*}[tbp]
\border
\vspace*{-5mm}
  \begin{center}
    \leavevmode
\begin{tabbing}
  If \= equal to \iz{deictic-day} (\engs{today, tomorrow, yesterday})
  or \engs{tonight}\\
  \> or \=(premodified by one of the deictic terms \eng{this, that, last,
    next} \\
  \> \>or one of the \iz{deictic-day} nouns \\
  \> \>or a quantifier such as \engs{every, some}) \\
  \> or post modified by \eng{ago, later} \\
  \> \into\  adverbial is a noun phrase (no preposition)\\
  \+ else create a prepositional phrase headed by:\\
  If  head is (\iz{hour} or \eng{night} and not modified) or
  \eng{beginning, end}:\\
  \>\into\  \eng{at}\\
  Else-If head is  \iz{day} or (\iz{period-of-day} and modified) \\
  \> \into\  \eng{on}\\
  Else \\
  \> \into\ \eng{in}
\end{tabbing}
\vspace*{-8mm}
  \end{center}
  \caption{Preposition choice for temporal adverbials}
  \label{fig:prep}
\vspace*{5mm}
\border
\end{figure*}

\eng{on} is optional for modified \iz{period-of-day} nouns like
\eng{Friday morning} in at least one author's idiolect, and for
\iz{day-of-week} nouns in general in American English.  In addition to
dialect the choice depends on domain and genre (stock reports and speech
tend to omit \eng{on}).  Depending on the users' requirements, some finer
gradations may need to be made.  By default we always generate \eng{on}
when it is considered optional, under the assumption that it is easier
to delete it afterwards than put it in.

\subsection{Deictic Concerns}

There is another concern for a practical machine translation system,
and that is the different choice of temporal expressions in different
cultures.  To take a well defined example, there is a marked
difference between temporal expressions in Japanese and English
on-line stock market reports in the Nikkei Biz Database.

To give three examples: in the Japanese reports, days are given as
days of the month while in English they are given as days of the week
\rsent{date-day}; expressions such as \jpn{week end}{sh\=umatsu} are
also translated as week days \rsent{week-day}; and periods within the
day are anchored to week day's in English but not in Japanese
\rsent{period}.

\begin{exe}
  \ex  \label{sent:date-day}
  \gll  \ldots beikoku-jikan-no \ul{12-13-nichi}-ni hirakareru FOMC \ldots\\
  {American time}-{\sc gen} 12-13-day-{on/at/in} hold FOMC\\
  \trans \ldots the U.S. FOMC meeting \ul{Tuesday and Wednesday} \ldots
  \ex \label{sent:week-day}
  \gll \ul{zen-sh\=umatsu}-no shikago nikkei-heikin-sakimono-daka \\
  {last week end}-{\sc gen} Chicago {Nikkei average futures high} \\
  \trans  Nikkei 225 futures gained in  Chicago \ul{last Friday}
  \ex  \label{sent:period} 
  \gll daisho-sh\=usei-wa \ul{maebike}-ni kake yasuneken-de momi-au \\
  {Osaka Stock Exchange}-{\sc top} {morning close}-{\sc dat} about
  {low range at} struggle \\
  \trans \ul{Monday morning trading} was confined to a boxed range at
  slightly lower levels
\end{exe}

For a machine translation system to be useful in such a domain it must
be able to make these conversions.  The conversion of dates to days is
relatively simple and could theoretically be done as part of a
separate pre or post editing process.  Changing stock market jargon to
mornings and afternoons requires knowledge of the domain, while
converting week beginnings and ends to days requires not just knowledge
of the actual date, but also on which day the stock market was open,
as, for example, the last trading day in the week is not necessarily a
Friday (Friday may have been a holiday).

We have added a special, domain-triggered section to our temporal
expression transfer module that converts dates to days, and
appropriately generates period of day expressions.  It uses common
date conversion routines, combined with the header date and time
information available in the on-line market reports themselves.  We
are in the process of extending the module to handle market opening
and closing date information.

Extensions like these are a small step into translation based on
domain understanding, which we hope will be the start of a long, but
interesting journey.

\section{Conclusion}
\label{sec:conc}

In this paper we present an outline of rules for translating Japanese
temporal phrases to English.  Because the rules only apply to a specific
limited class of input, and there are many lexical idiosyncrasies, we
argue that these rules are better thought of as a separate module,
although integrated with the entire system.  We also emphasize the need
to consider the context of temporal expressions, giving the example of
Japanese and English Stock market reports, where knowledge of the date
the report was prepared is essential to translate it.

\subsection*{Acknowledgements}  The authors thank Tim Baldwin,
Rodney Huddleston, Yukie Kuribayashi, Roland Sussex and the members of
the NTT Machine Translation Research Group for their disscussion and comments;
as well as Satoshi Mizuno and Shinsuke Okuyama of NTT Software for
their help in the implementation.

\bibliographystyle{acl}


\end{document}